\documentclass{article}

\usepackage{arxiv}
\usepackage[utf8]{inputenc} 
\usepackage[T1]{fontenc}    
\usepackage{hyperref}       
\usepackage{url}            
\usepackage{booktabs}       
\usepackage{amsfonts}       
\usepackage{nicefrac}       
\usepackage{microtype}      
\usepackage{lipsum}
\usepackage{graphicx}
\usepackage{multicol}

\title{Generaci\'on autom\'atica de frases literarias en espa\~nol}

\author{
  Luis-Gil Moreno-Jiménez\\
  Université d'Avignon/LIA\\Universidad Tecnol\'ogica de la Selva\\
  \texttt{luis-gil.moreno-jimenez}\\
  \texttt{@alumni.univ-avignon.fr}
  \And 
  Juan-Manuel Torres-Moreno\\
  Université d'Avignon/LIA\\Polytechnique Montréal\\
  \texttt{juan-manuel.torres}\\
  \texttt{@univ-avignon.fr}\\
  \And
  Roseli S. Wedemann\\
  Universidade do Estado do Rio de Janeiro\\
  \texttt{roseli@ime.uerj.br}\\
}

\begin{document} 

\maketitle

\begin{abstract}
  En este trabajo presentamos un estado del arte en el área de la Creatividad Computacional (CC). 
  En particular abordamos la generaci\'on automática de frases literarias en espa\~nol. 
  Proponemos tres modelos de generación textual basados principalmente en algoritmos estadísticos y an\'alisis sint\'actico superficial. 
  Presentamos también algunos resultados preliminares bastante alentadores.
\end{abstract}

\section{Introducción}

Los investigadores en Procesamiento de Lenguaje Natural (PLN) durante mucho tiempo han utilizado corpus constituidos por documentos enciclopédicos (notablemente Wikipedia), period\'isticos (peri\'odicos o revistas) o especializados (documentos legales, cient\'ificos o técnicos) para el desarrollo y pruebas de sus modelos \cite{torres2014,cunhaCSSMV11,sierra}.

La utilizaci\'on y estudios de corpora literarios sistem\'aticamente han sido dejados a un lado por varias razones. 
En primer lugar, el nivel de discurso literario es m\'as complejo que los otros géneros. 
En segundo lugar, a menudo, los documentos literarios hacen referencia a mundos o situaciones imaginarias o aleg\'oricas, a diferencia de los otros géneros que describen sobre todo situaciones o hechos factuales. 
Estas y otras características presentes en los textos literarios, vuelven sumamente compleja la tarea de an\'alisis automático de este tipo de textos.
En este trabajo nos proponemos utilizar corpora literarios, a fin de generar realizaciones literarias (frases nuevas) no presentes en dichos corpora.

La producción de textos literarios es el resultado de un proceso donde una persona hace uso de aptitudes creativas.
Este proceso, denominado ``proceso creativo'', ha sido analizado por \cite{boden2004creative}, quien propone tres tipos básicos de creatividad: 
la primera, Creatividad Combinatoria (CCO), donde se fusionan elementos conocidos para la generación de nuevos elementos.
La segunda, Creatividad Exploratoria (CE), donde la generación ocurre a partir de la observación o exploración. 
La tercera, Creatividad Transformacional (CT), donde los elementos generados son producto de alteraciones o experimentaciones aplicadas al dominio de la CE.

Sin embargo, cuando se pretende automatizar el proceso creativo, la tarea debe ser adaptada a métodos formales que puedan ser realizados en un algoritmo. 
Este proceso automatizado da lugar a un nuevo concepto denominado Creatividad Computacional (CC), introducido por \cite{perez2015creatividad}, quien retoma para ello la CT y la CE propuestas por \cite{boden2004creative}.

La definici\'on de literatura no tiene un consenso universal, y muchas variantes de la definición pueden ser encontradas. 
En este trabajo optaremos por introducir una definici\'on pragm\'atica de frase literaria, que servir\'a para nuestros modelos y experimentos. 

\textit{Definici\'on. Una frase literaria es una frase que se diferencia de las frases en lengua general, porque contiene elementos (nombres, verbos,  adjetivos, adverbios) que son percibidos como elegantes o menos coloquiales que sus equivalentes en lengua general.} 




En particular, proponemos crear artificialmente frases literarias utilizando modelos generativos y aproximaciones sem\'anticas basados en corpus de lengua literaria.
La combinaci\'on de esos modelos da lugar a una homosintaxis, es decir, la producci\'on de texto nuevo a partir de formas de discurso de diversos autores.
La homosintaxis no tiene el mismo contenido sem\'antico, ni siquiera las mismas palabras, aunque guarda la misma estructura sintáctica.

En este trabajo proponemos estudiar el problema de la generaci\'on de texto literario original en forma de frases aisladas, no a nivel de p\'arrafos. 
La generaci\'on de p\'arrafos puede ser objeto de trabajos futuros. 
Una evaluación de la calidad de las frases generadas por nuestro sistema ser\'a presentada.

Este art\'iculo est\'a estructurado como sigue. 
En la Secci\'on \ref{sec:arte} presentamos un estado del arte de la creatividad computacional. 
En la Secci\'on \ref{sec:corpus} describimos los corpus utilizados. 
Nuestros modelos son descritos en la Secci\'on \ref{sec:modelos}. 
Los resultados y su interpretaci\'on se encuentran en la Secci\'on \ref{sec:resultados}. 
Finalmente la Secci\'on \ref{sec:conclusiones} presenta algunas ideas de trabajos futuros antes de concluir.

\section{
Trabajos previos}
\label{sec:arte} 

La generaci\'on de texto es una tarea relativamente cl\'asica, que ha sido estudiada en diversos trabajos. Por ejemplo, 
\cite{Szymaski2002HiddenMM} presentan un modelo basado en cadenas de Markov para la generaci\'on de texto en idioma polaco. 
Los autores definen un conjunto de estados actuales y calculan la probabilidad de pasar al  estado siguiente. 
La ecuación (\ref{eq:probhmm}) calcula la probabilidad de pasar al estado $X_{i}$ a partir de $X_{j}$,
\begin{equation}
\label{eq:probhmm}
P_{ij}(X_{i}|X_{j}) = P(X_{i} \cap X_{j})|P(X_{j})\, .
\end{equation}
Para ello, se utiliza una matriz de transici\'on, la cual contiene las probabilidades de transición de un estado actual $X_i$ a los posibles estados futuros $X_{i+1}$. 
Cada estado puede estar definido por $n$-gramas de letras o de palabras.

La tarea inicia en un estado $X_i$ dado por el usuario.
Posteriormente, usando la matriz de transici\'on, se calcula la probabilidad de pasar al estado siguiente $X_{i+1}$. 
En ese momento el estado predicho $X_{i+1}$ se convierte en el estado actual $X_i$, repitiendo este proceso hasta satisfacer una condición. 
Este método tiene un buen comportamiento al generar palabras de 4 o 5 letras. 
En polaco esta longitud corresponde a la longitud media de la mayor parte de las palabras \cite{ultrastemming}. 

También hay trabajos que realizan an\'alisis m\'as profundos para generar no solamente palabras, sino p\'arrafos completos. %
\cite{Sridhara:2010:TAG:1858996.1859006} presentan un algoritmo que genera automáticamente comentarios descriptivos para bloques de código (métodos) en Java.
Para ello, se toma el nombre del método y se usa como la acci\'on o idea central de la descripción a generar. 
Posteriormente se usan un conjunto de heurísticas, para seleccionar las líneas de código del método que puedan aportar mayor información, y se procesan para generar la descripción.
La tarea consiste en construir sintagmas, a partir de la idea central dada por el nombre del método, y enriquecerlos con la información de los elementos extraídos.
Por ejemplo, si hay un método \texttt{removeWall(Wall x)} y se encuentra la llamada al método \texttt{removeWall(oldWall)}, la descripci\'on generada podría ser: ``Remove old Wall''.
Obteniéndose la acción (verbo) y el objeto (sustantivo) directamente del nombre del método y el adjetivo a partir de la llamada. 
Estas ideas permiten a los autores la generación de comentarios extensos sin perder la coherencia y la gramaticalidad.

También se encuentran trabajos de generación textual que
se proponen como meta resultados con un valor más artístico.
\cite{Riedl2006} presentan un conjunto de algoritmos para la generación de una guía narrativa basada en la idea de Creatividad Exploratoria \cite{boden2004creative}.
El modelo establece i/ un conjunto universal U de conceptos relevantes relacionados a un dominio; ii/ un modelo generador de texto; iii/ un subconjunto de conceptos S que pertenecen al conjunto universal U; y iv/ algoritmos encargados de establecer las relaciones entre U y S para generar nuevos conceptos. 
Estos nuevos conceptos serán posteriormente comparados con los conceptos ya existentes en U para verificar la coherencia y relación con la idea principal.
Si los resultados son adecuados, estos nuevos conceptos se utilizan para dar continuación a la narrativa. 

Son diversos los trabajos que están orientados a la generación de una narrativa ficticia como cuentos o historias. 
\cite{clark-etal-2018-neural} proponen un modelo de generación de texto narrativo a partir del análisis de \textit{entidades}. 
Dichas \textit{entidades} son palabras (verbos, sustantivos o adjetivos) dentro de un texto que serán usados para generar la frase siguiente. 
El modelo recupera las \textit{entidades} obtenidas de tres fuentes principales: la frase actual, la frase previa y el documento completo (contexto), y las procesa con una red neuronal para seleccionar las mejores de acuerdo a diversos criterios.
A partir de un conjunto de heurísticas, se analizaron las frases generadas para separar aquellas que expresaran una misma idea (paráfrasis), de aquellas que tuvieran una relación entre sus \textit{entidades} pero con ideas diferentes.

La generaci\'on de texto literario es un proceso muy diferente a la generación de texto aleatorio \cite{lebret2016neural,welleck2019non} y tampoco se limita a una idea o concepto general. 
El texto literario est\'a destinado a ser un documento elegante y agradable a la lectura, haciendo uso de figuras literarias y un vocabulario distinto al empleado en la lengua general.
Esto da a la obra una autenticidad y define el estilo del autor.
El texto literario también debe diferenciarse de las estructuras rígidas o estereotipadas de los géneros periodístico, enciclopédico o científico.

\cite{Zhang2014Poetry} proponen un modelo para la generación de poemas y se basa en dos premisas básicas: \textsl{¿qué decir?} y \textsl{¿cómo decirlo?}
La propuesta parte de la selección de un conjunto de frases tomando como guía una lista de palabras dadas por el usuario.
Las frases son procesadas por un modelo de red neuronal \cite{Mikolov2012Context}, para construir combinaciones coherentes y formular un contexto. 
Este contexto es analizado para identificar sus principales elementos y generar las l\'ineas del poema, que también pasarán a formar parte del contexto.
El modelo fue evaluado manualmente por 30 expertos en una escala de 1 a 5, analizando legibilidad, coherencia y significatividad en frases de 5 palabras, obteniendo una  precisión de 0.75.
Sin embargo, la coherencia entre frases result\'o ser muy pobre.

\cite{Oliveira2012,Oliveira2015} proponen  un modelo de generación de poemas a base de plantillas. 
El algoritmo inicia con un conjunto de frases relacionadas a partir de palabras clave.
Las palabras clave sirven para generar un contexto.
Las frases son procesadas usando el sistema PEN\footnote{Disponible en: \url{http://code.google.com/p/pen}} para obtener su información gramatical. 
Esta información es empleada para la generación de nuevas platillas gramaticales y finalmente la construcción de las líneas del poema, tratando de mantener la coherencia y la gramaticalidad.

El modelo sentiGAN \cite{wang2018sentigan} pretende generar texto con un contexto emocional. 
Se trata de una actualización del modelo GAN  (\textsl{Generative Adversarial Net}) \cite{goodfellow2014generative} que ha producido resultados alentadores en la generación textual, aunque con ciertos problemas de calidad y coherencia. 
Se utiliza el análisis semántico de una entrada proporcionada por el usuario que sirve para la creación del contexto.
La propuesta principal de SentiGAN sugiere establecer un número definido de generadores textuales que deberán producir texto relacionado a una emoción definida.
Los generadores son entrenados bajo dos esquemas: i/ una serie de elementos lingüísticos que deben ser evitados para la generación del texto; y ii/ un conjunto de elementos relacionados con la emoción ligada al generador.
A través de cálculos de distancia, heurísticas y modelos probabilísticos, el generador crea un texto lo más alejado del primer esquema y lo más cercano al segundo.

También existen trabajos con un alcance más corto pero de mayor precisión. 
\cite{huang2012improving} proponen la evaluación de un conjunto de datos con un modelo basado en redes neuronales para la generación de subconjuntos de multi-palabras. 
Este mismo análisis, se considera en \cite{fu2014learning}, en donde se busca establecer o detectar la relación hiperónimo-hip\'onimo con la ayuda del modelo de \textit{Deep Learning} Word2vec  \cite{mikolov2013linguistic}.
La propuesta de \cite{fu2014learning} reporta una precisión de 0.70 al ser evaluado sobre un corpus manualmente etiquetado.

La literatura es una actividad artística que exige capacidades creativas importantes y que ha llamado la atención de científicos desde hace cierto tiempo.
\cite{perez2015creatividad} realiza un estado del arte interesante donde menciona algunos trabajos que tuvieron un primer acercamiento a la obra literaria desde una perspectiva superficial. 
Por ejemplo, el modelo ``Through the park''~\cite{montfort2008b}, es capaz de generar narraciones históricas empleando la \textrm{elipsis}. 
Esta técnica es empleada para manipular, entre otras cosas, el ritmo de la narraci\'on.
En los trabajos ``About So Many Things'' \cite{montfort2008c} y  ``Taroko Gorge'' \cite{montfort2009} se muestran textos generados autom\'aticamente.
El primero de ellos genera estrofas de 4 líneas estrechamente relacionadas entre ellas.
Eso se logra a través de un análisis gramatical que establece conexiones entre entidades de distintas líneas.
El segundo trabajo muestra algunos poemas cortos generados autom\'aticamente con una estructura m\'as compleja que la de las estrofas.
El inconveniente de ambos enfoques es el uso de una estructura inflexible, lo que genera textos repetitivos con una gramaticalidad limitada.

El proyecto MEXICA modela la generación colaborativa de narraciones \cite{perez2015creatividad}. 
El prop\'osito es la generación de narraciones completas utilizando obras de la época Precolombina. 
MEXICA genera narraciones simulando el proceso creativo de E-R \textit{(Engaged y Reflexive)} \cite{sharpies1999we}.
Este proceso se describe como la acción, donde el autor trae a su mente un conjunto de ideas y contextos y establece una conexión coherente entre estas (E).
Posteriormente se reflexiona sobre las conexiones establecidas y se evalúa el resultado final para considerar si este realmente satisface lo esperado (R).
El proceso itera hasta que el autor lo considera concluido. 

\section{Corpus utilizados}
\label{sec:corpus}


\subsection{Corpus 5KL}
\label{corpus5KL}

Este corpus fue constituido con aproximadamente 5~000 documentos (en su mayor parte libros) en español. 
Los documentos originales, en formatos heterogéneos, fueron procesados para crear un \'unico documento codificado en \textit{utf8}.
Las frases fueron segmentadas autom\'aticamente, usando un programa en PERL 5.0 y expresiones regulares, para obtener una frase por l\'inea.

Las caracter\'isticas del corpus 5KL se encuentran en la Tabla~\ref{tab:1}. 
Este corpus es empleado para el entrenamiento de los modelos de aprendizaje profundo (\textit{Deep Learning}, Secci\'on \ref{sec:modelos}).

\begin{table}[htb]
  \centering
  \begin{tabular}{l|rrr}
    \toprule
   & \textbf{Frases} & \textbf{Palabras} & \textbf{Caracteres} \\
    \midrule
   \textbf{5KL}   & 9 M & 149 M & 893 M \\ \hline
    Media por  &   &   &    \\
    documento & 2.4 K & 37.3 K & 223 K \\
    \bottomrule
  \end{tabular}
  \caption{Corpus 5KL compuesto de 4 839 obras literarias.}
  \label{tab:1}
\end{table}

El corpus literario 5KL posee la ventaja de ser muy extenso y adecuado para el aprendizaje autom\'atico.
Tiene sin embargo, la desventaja de que no todas las frases son \textsl{necesariamente} ``frases literarias''.
Muchas de ellas son frases de lengua general: 
estas frases a menudo otorgan una fluidez a la lectura y proporcionan los enlaces necesarios a las ideas expresadas en las frases literarias.

Otra desventaja de este corpus es el ruido que contiene. 
El proceso de segmentaci\'on puede producir errores en la detecci\'on de  fronteras de frases.
También los n\'umeros de p\'agina, cap\'itulos, secciones o \'indices producen errores.
No se realiz\'o ning\'un proceso manual de verificaci\'on, por lo que a veces se introducen informaciones indeseables: \textsl{copyrights}, datos de la edici\'on u otros.
Estas son, sin embargo, las condiciones que presenta un corpus literario real.

\subsection{Corpus 8KF}
\label{corpus8K}


Un corpus heterog\'eneo de casi 8~000 frases literarias fue constituido manualmente a partir de poemas, discursos, citas, cuentos y otras obras. 
Se evitaron cuidadosamente las frases de lengua general, y también aquellas demasiado cortas ($N \le 3$ palabras) o demasiado largas ($N \ge 30$ palabras).
El vocabulario empleado es complejo y estético, además que el uso de ciertas figuras literarias como la rima, la anáfora, la metáfora y otras pueden ser observadas en estas frases. 

Las caracter\'isticas del corpus 8KF se muestran en la Tabla~\ref{tab:2}.
Este corpus fue utilizado principalmente en los dos modelos generativos: modelo basado en cadenas de Markov (Secci\'on \ref{sec:modelomarkov}) y modelo basado en la generación de \textrm{Texto enlatado} (\textit{Canned Text}, Secci\'on \ref{sec:modelolata}).

\begin{table}[htb]
  \centering
  \begin{tabular}{lrrr}
    \toprule
     & \textbf{Frases} & \textbf{Palabras} & \textbf{Caracteres} \\
    \midrule
    \textbf{8KF}    & 7 679 & 114 K & 652 K \\ \hline
    Media           &  &   &   \\                
    por frase & -- & 15 & 85 \\
    \bottomrule
  \end{tabular}
  \caption{Corpus 8KF compuesto de 7 679 frases literarias.}
  \label{tab:2}
\end{table}

\section{Modelos propuestos}
\label{sec:modelos}

En este trabajo proponemos tres modelos h\'ibridos (combinaciones de modelos generativos cl\'asicos y aproximaciones sem\'anticas) para la producción de frases literarias. 
Hemos adaptado dos modelos generativos, usando an\'alisis sint\'actico superficial (\textsl{shallow parsing}) y un modelo de aprendizaje profundo (\textsl{Deep Learning}) \cite{deeplearning}, combinados con tres modelos desarrollados de aproximación semántica.

En una primera fase, los modelos generativos recuperan la información gramatical de cada palabra del corpus 8KF (ver Sección \ref{sec:corpus}), en forma de etiquetas POS (\textsl{Part of Speech}), a través de un análisis morfosintáctico. 
Utilizamos Freeling \cite{freeling} que permite análisis lingüísticos en varios idiomas\footnote{
Puede ser obtenido en la direcci\'on: \url{http://nlp.lsi.upc.edu/freeling}}.
Por ejemplo, para la palabra ``Profesor'' Freeling genera la etiqueta POS \texttt{[NCMS000]}.
La primera letra indica un sustantivo (\textbf{N}oun), la segunda un sustantivo común (\textbf{C}ommon); la tercera indica el género masculino (\textbf{M}ale) y la cuarta da información de número (\textbf{S}ingular). 
Los 3 \'ultimos caracteres dan información detallada del campo semántico,  entidades nombradas, etc.\footnote{Más detalles de las etiquetas Freeling en \url{http://blade10.cs.upc.edu/freeling-old/doc/tagsets/tagset-es.html}}
En nuestro caso usaremos solamente los 4 primeros niveles de las etiquetas.

Con los resultados del análisis morfosintáctico, se genera una salida que llamaremos \textit{Estructura gramatical vac\'ia}  (EGV): 
compuesta exclusivamente de una secuencia de etiquetas POS; o \textit{Estructura gramatical parcialmente vac\'ia} (EGP), compuesta de etiquetas POS y de palabras funcionales (art\'iculos, pronombres, conjunciones, etc.).

En la segunda fase, las etiquetas POS (en la EGV y la EGP) ser\'an reemplazadas por un vocabulario adecuado usando ciertas aproximaciones sem\'anticas.

La producci\'on de una frase $f(Q,N)$ es guiada por dos par\'ametros: 
un contexto representado por un término $Q$ (o \textit{query}) y una longitud $3 \le N \le 15$, dados por el usuario. 
Los corpus 5KL y 8KF son utilizados en varias fases de la producci\'on de las frases $f$.

\begin{itemize} 
\item El Modelo 1 est\'a compuesto por: 
i/ un modelo generativo estoc\'astico basado en cadenas de Markov para la selección de la próxima etiqueta POS
usando el algoritmo de Viterbi; y ii/ un modelo de aprendizaje profundo (Word2vec), para recuperar el vocabulario que reemplazará la secuencia de etiquetas POS.

\item El Modelo 2 es una combinación de:  i/ el modelo generativo de \textrm{Texto enlatado}; y ii/ un modelo Word2vec, con un c\'alculo de distancias entre diversos vocabularios que han sido constituidos manualmente.

\item El Modelo 3 utiliza: i/ la generaci\'on de \textrm{Texto enlatado}; y ii/ una interpretaci\'on geométrica del aprendizaje profundo. 
Esta interpretaci\'on est\'a basada en una b\'usqueda de informaci\'on iterativa (\textit{Information Retrieval}, IR), que realiza simult\'aneamente un alejamiento de la sem\'antica original y un acercamiento al \textit{query} $Q$ del usuario.
\end{itemize}

\begin{figure}[h]
\centering
\includegraphics[width=9cm]{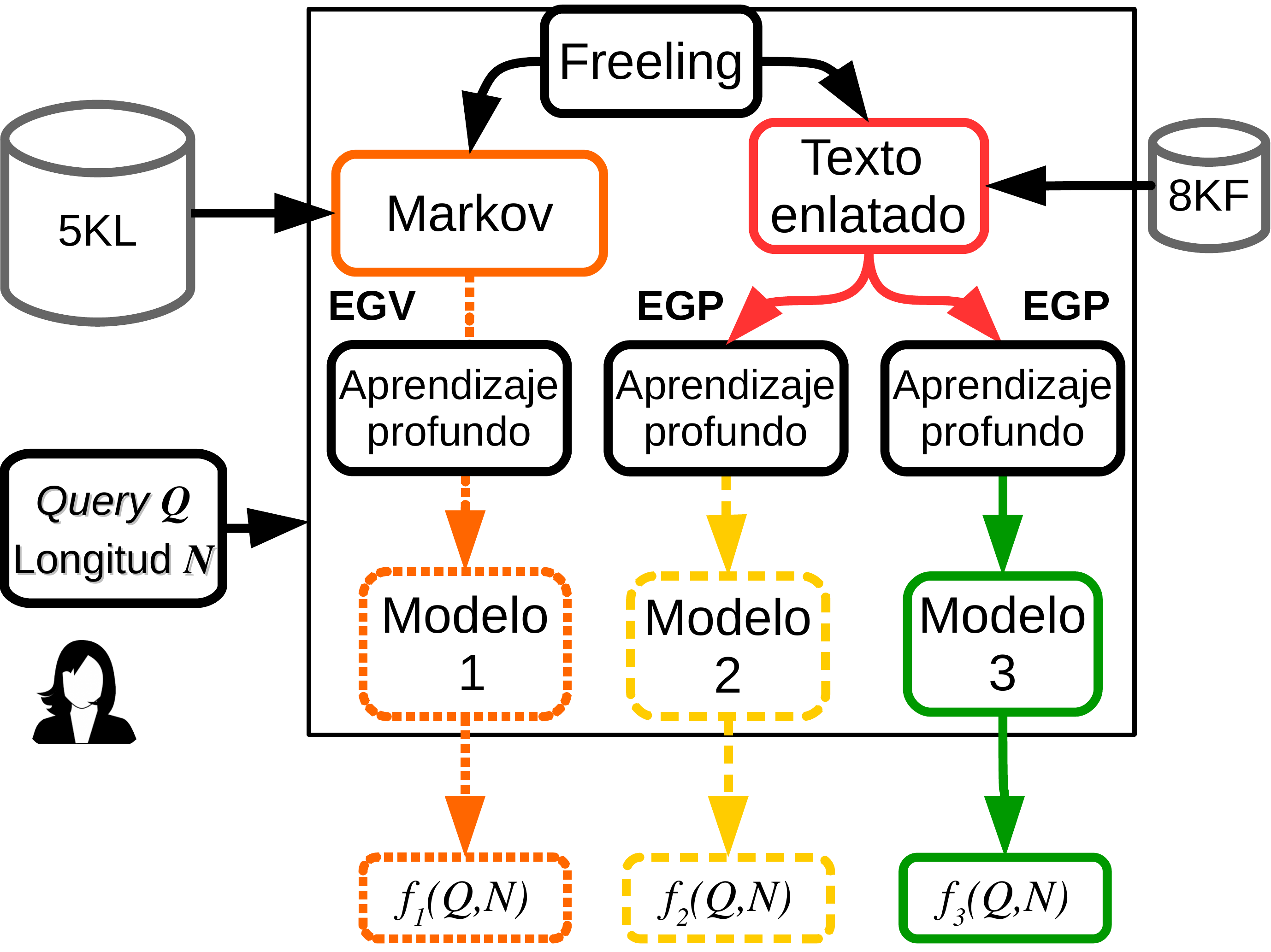}
\caption{Arquitectura general de los modelos.}
\label{fig:modelos}
\end{figure}


\subsection{Modelo generativo estoc\'astico usando cadenas de Markov}
\label{sec:modelomarkov}

Este modelo generativo, que llamaremos \textit{Modelo de Markov\/}, está basado en el algoritmo de Viterbi y las cadenas de Markov \cite{manning99foundations}, donde se selecciona una etiqueta POS con la máxima probabilidad de ocurrencia, para ser agregada al final de la secuencia actual.

Utilizamos el corpus de frases literarias 8KF (ver Secci\'on \ref{corpus8K}), que fue convenientemente filtrado para eliminar  \textsl{tokens} indeseables: números, siglas, horas y fechas.
El corpus filtrado se analiz\'o usando Freeling,
que recibe en entrada una cadena de texto y entrega el texto con una etiqueta POS para cada palabra. 
El corpus es analizado frase a frase, reemplazando cada palabra por su respectiva etiqueta POS. 
Al final del an\'alisis, se obtiene un nuevo corpus 8KPOS con $s = 7~679$ secuencias de etiquetas POS, correspondientes al mismo n\'umero de frases del corpus 8KF. 
Las secuencias del corpus 8KPOS sirven como conjunto de entrenamiento para el algoritmo de Viterbi, que calcula las probabilidades de transición, que serán usadas para generar cadenas de Markov.

Las $s$ estructuras del corpus 8KPOS procesadas con el algoritmo de Viterbi son representadas en una matriz de transición $P_{[s \times  s]}$.
$P$ ser\'a utilizada para crear nuevas secuencias de etiquetas POS no existentes en el corpus 8KPOS, simulando un proceso creativo. 
Nosotros hemos propuesto el algoritmo \textit{Creativo-Markov} que describe este procedimiento.

En este algoritmo, $X_i$ representa el estado de una etapa de la creación de una frase, en el instante $i$, que corresponde a una secuencia de etiquetas POS. 
Siguiendo un procedimiento de Markov, en un instante $i$ se selecciona la próxima etiqueta POS$_{i+1}$, 
con máxima probabilidad de ocurrencia, dada la última etiqueta POS$_i$ de la secuencia $X_{i}$. 
La etiqueta POS$_{i+1}$ será agregada al final de $X_{i}$ para generar el estado $X_{i+1}$.
$P(X_{i+1}=Y|X_{i}=Z)$
es la probabilidad de transici\'on de un estado a otro, obtenido con el algoritmo de Viterbi.
Se repiten las transiciones, hasta alcanzar una longitud deseada.

\begin{figure}[h]
\centering
\includegraphics[width=9cm]{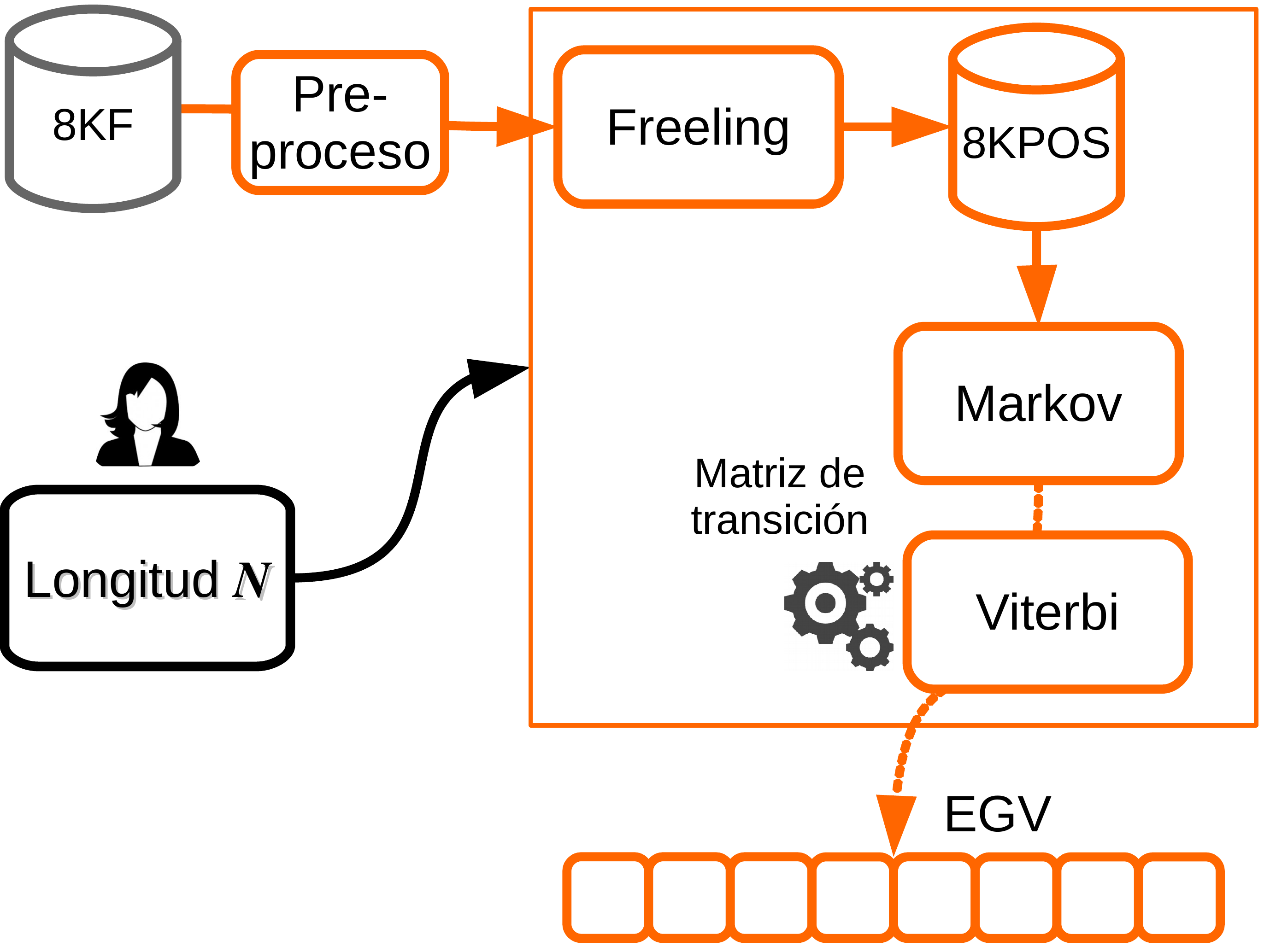}
\caption{Modelo generativo estoc\'astico (Markov) que produce una estructura gramatical vac\'ia EGV.}
\label{fig:markov}
\end{figure}

El resultado es una EGV, donde cada cuadro vac\'io representa una etiqueta POS que ser\'a remplazada por una palabra en la etapa final de generaci\'on de la nueva frase.
El remplazo se realiza usando un modelo de aprendizaje profundo (Secci\'on \ref{modeloword2vect-1}).
La arquitectura general de este modelo se muestra en la Figura \ref{fig:markov}. 

\subsection{Modelo generativo basado en \textrm{Texto enlatado}}
\label{sec:modelolata}

El algoritmo \textit{creativo-Markov} del \textit{Modelo de Markov} logra reproducir patrones lingüísticos (secuencias POS) detectados en el corpus 8KPOS, pero de corta longitud.
Cuando se intent\'o extender la longitud de las frases a $N>6$ palabras, no fue posible mantener la coherencia y legibilidad (como se ver\'a en la Sección \ref{modeloword2vect-1}).
Decidimos entonces utilizar métodos de generación textual guiados por estructuras morfosintácticas fijas: el \textrm{Texto enlatado}. 
\cite{molins2015jsrealb} argumentan que el uso de estas estructuras ahorran tiempo de análisis sintáctico y permite concentrarse directamente en el vocabulario.

La técnica de \textrm{Texto enlatado} ha sido empleada también en varios trabajos, con objetivos específicos. 
\cite{Templatebasedaumented, Templatebased} desarrollaron modelos para la generación de diálogos y frases simples. 
Esta técnica es llamada ``Generación basada en plantillas'' (\textit{Template-based Generation}) o de manera intuitiva, \textrm{Texto enlatado}\footnote{\url{ http://projects.ict.usc.edu/nld/cs599s13/LectureNotes/cs599s13dialogue2-13-13.pdf}}.

Decidimos emplear \textrm{texto enlatado} para la generación textual usando un corpus de plantillas (\textit{templates}), construido a partir del corpus 8KF (Sección \ref{sec:corpus}).
Este corpus contiene estructuras gramaticales flexibles que pueden ser manipuladas para crear nuevas frases. 
Estas plantillas pueden ser seleccionadas aleatoriamente o a través de heurísticas, seg\'un un objetivo predefinido.

Una plantilla es construida a partir de las palabras de una frase $f$, donde se reemplazan \'unicamente las palabras llenas de las clases verbo, sustantivo o adjetivo $\{ V, S, A \}$, por sus respectivas etiquetas POS. 
Las otras palabras, en particular las palabras funcionales, son conservadas.
Esto producir\'a una \textit{estructura gramatical parcialmente vac\'ia, EGP}.
Posteriormente las etiquetas podrán ser reemplazadas por palabras (términos), relacionadas con el contexto definido por el \textit{query} $Q$ del usuario.

El proceso inicia con la selección aleatoria de una frase original $f_{o} \in $ corpus 8KF de longitud $|f_{o}|=N$. 
$f_{o}$ será analizada con Freeling para identificar los sintagmas. 
Los elementos $\{ V, S, A \}$ de los sintagmas de $f_{o}$ ser\'an reemplazados por sus respectivas etiquetas POS.
Estos elementos  son los que mayor información aportan en cualquier texto, independientemente de su longitud o género \cite{bracewell2005multilingual}.
Nuestra hip\'otesis es que al cambiar solamente estos elementos, simulamos la generaci\'on de frases por homosintaxis: sem\'antica diferente, misma estructura\footnote{Al contrario de la paráfrasis que busca conservar completamente la sem\'antica, alterando completamente la estructura sintáctica.}.

\begin{figure}[h]
\centering
\includegraphics[width=9cm]{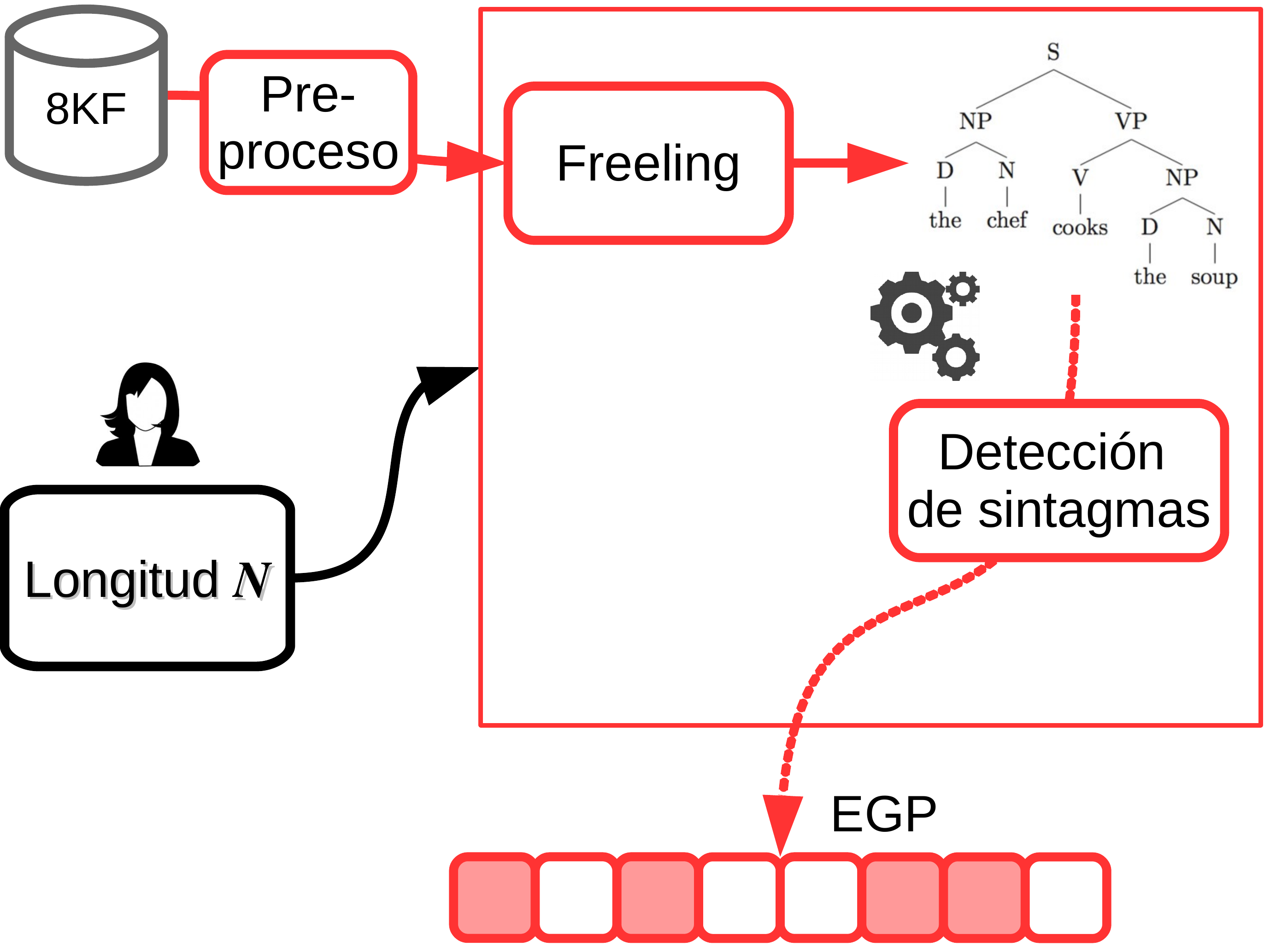}
\caption{Modelo generativo de \textrm{Texto enlatado} que produce una estructura parcialmente vac\'ia.}
\label{fig:textinlata}
\end{figure}

La salida de este proceso es una estructura híbrida parcialmente vac\'ia (EGP) con palabras funcionales que dan un soporte gramatical y las etiquetas POS.
La arquitectura general de este modelo se ilustra en la Figura \ref{fig:textinlata}. 
Los cuadros llenos representan palabras funcionales y los cuadros vac\'ios etiquetas POS a ser reemplazadas.

\subsection{Modelo 1: Markov y aprendizaje profundo}
\label{modeloword2vect-1}

Los modelos generativos
generan estructuras gramaticales vac\'ias (EGV) o parcialmente vac\'ias (EGP)
que pueden ser manipuladas para generar nuevas frases $f(Q,N)$.
La idea es que las frases $f$ sean generadas por homosintaxis. 
En esta sección, proponemos un modelo de aproximaci\'on sem\'antica que utiliza el algoritmo Word2vec (basado en aprendizaje profundo), combinado con el modelo generativo de Markov descrito en la Secci\'on \ref{sec:modelomarkov}. 
El proceso se describe a continuación.

El corpus 5KL es pre-procesado para uniformizar el formato del texto, eliminando caracteres que no son importantes para el análisis semántico: 
puntuación, números, etc.
Esta etapa prepara los datos de entrenamiento del algoritmo de aprendizaje profundo que utiliza una representación vectorial del corpus 5KL.
Para el aprendizaje profundo utilizamos la biblioteca Gensim\footnote{Disponible en: \url{https://pypi.org/project/gensim/}}, la versión en Python de Word2vec\footnote{\url{https://towardsdatascience.com/introduction-to-word-embedding-and-word2vec-652d0c2060fa}}.
Con este algoritmo se obtiene un conjunto de palabras asociadas (\textit{embeddings}) a un contexto definido por un \textit{query} $Q$.
Word2vec recibe un término $Q$ y devuelve un léxico $L(Q)=(w_1,w_2,...,w_m)$ que representa un conjunto de $m$ palabras semánticamente próximas a $Q$.
Formalmente, Word2vec: $Q \rightarrow L(Q)$.

El próximo paso consiste en procesar la EGV producida por Markov.
Las etiquetas POS ser\'an identificadas y clasificadas como 
POS$_{\Phi}$ funcionales (correspondientes a puntuaci\'on y \textrm{palabras funcionales}) y POS$_\lambda$ llenas $\in \{ V, S, A \}$ (verbos, sustantivos, adjetivos). 

Las etiquetas POS$_\Phi$ ser\'an reemplazadas por palabras obtenidas de recursos lingü\'isticos (diccionarios) constru\'idos con la ayuda de Freeling. 
Los diccionarios
consisten en entradas de pares: POS$_\Phi$ y una lista de palabras y signos asociados, formalmente POS$_\Phi$ $\rightarrow$ $l$(POS$_\Phi)=(l_1,l_2,...,l_j)$.
Se reemplaza aleatoriamente cada POS$_\Phi$ por una palabra de $l$ que corresponda a la misma clase gramatical.

Las etiquetas POS$_\lambda$ serán reemplazadas por las palabras producidas por Word2vec $L(Q)$.
Si ninguna de las palabras de $L(Q)$ tiene la forma sintáctica exigida por POS$_\lambda$, empleamos la biblioteca PATTERN\footnote{\url{https://www.clips.uantwerpen.be/pattern}} para realizar conjugaciones o conversiones de género y/o número y reemplazar correctamente POS$_\lambda$.

\begin{figure}[h]
\centering
\includegraphics[width=9cm]{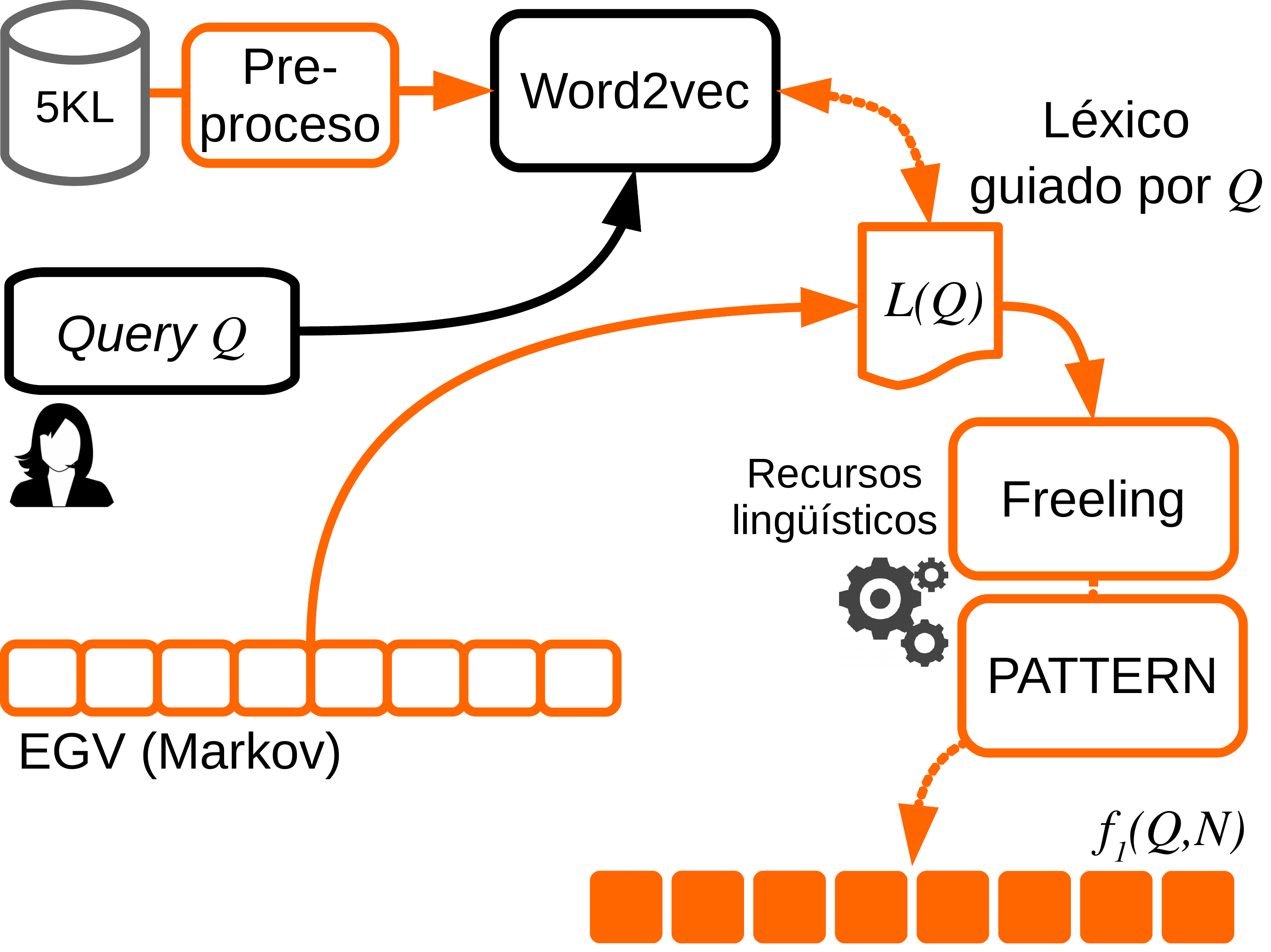}
\caption{Modelo 1: Aproximación semántica usando Markov y aprendizaje profundo.
}
\label{fig:ms1}
\end{figure}

Si el conjunto de palabras $L(Q)$, no contiene ning\'un tipo de palabra llena, que sea adecuada o que pueda manipularse con la biblioteca PATTERN, para reemplazar las etiquetas POS$_\lambda$, se toma otra palabra, $w_i \in L(Q)$, lo más cercana a $Q$ (en funci\'on de la distancia producida por Word2vec).
Se define un nuevo $Q*=w_i$ que será utilizado para generar un nuevo conjunto de palabras $L(Q*)$. 
Este procedimiento se repite hasta que $L(Q*)$ contenga una palabra que pueda reemplazar la POS$_{\lambda}$ en cuesti\'on. 
El resultado de este procedimiento es una nueva frase $f$ que no existe en los corpora 5KL y 8KF. 
La Figura \ref{fig:ms1} muestra el proceso descrito.

\subsection{Modelo 2: \textrm{Texto enlatado}, aprendizaje profundo y an\'alisis morfosintáctico}
\label{modeloword2vect-2}

En este modelo proponemos una combinación entre el modelo de \textrm{Texto enlatado} (Sección \ref{sec:modelolata}) y un algoritmo de aprendizaje profundo con Word2vec entrenado sobre el corpus 5KL.
El objetivo es eliminar las iteraciones del Modelo 1, que son necesarias cuando las etiquetas POS\footnote{Por motivos de claridad de la notaci\'on, en esta secci\'on y en la siguiente una etiqueta POS$_{\lambda}$ ser\'a designada solamente por POS.} no pueden ser reemplazadas con el léxico $L(Q)$. 

Se efect\'ua un análisis morfosintáctico del corpus 5KL usando Freeling y se usan las etiquetas POS para crear conjuntos de palabras que posean la misma informaci\'on gramatical (etiquetas POS idénticas).
Una Tabla Asociativa (TA) es generada como resultado de este proceso. 
La TA consiste en $k$ entradas de pares POS$_k$ y una lista de palabras asociadas. 
Formalmente POS$_k \rightarrow V_k =\{v_{k,1},v_{k,2},...,v_{k,i}\}$.

El Modelo 2 es ejecutado una sola vez para cada etiqueta POS$_k$. 
La EGP no será reemplazada completamente: las palabras funcionales y los signos de puntuación son conservados.

Para generar una nueva frase se reemplaza cada etiqueta POS$_k \in$ EGP, $k=1,2,...$,  por una palabra adecuada. 
Para cada etiqueta POS$_k$, se recupera el léxico $V_k$ a partir de TA. 

El vocabulario es procesado por el algoritmo Word2vec, que calcula el valor de proximidad (distancia) entre cada palabra del vocabulario $v_{k,i}$ y el \textit{query} $Q$ del usuario, $dist(Q,v_{k,i})$. 
Después se ordena el vocabulario $V_k$ en forma descendente seg\'un los valores de proximidad $dist(Q,v_{k,i})$ y se escoge aleatoriamente uno de los primeros tres elementos para reemplazar la etiqueta POS$_k$ de la EGP.

\begin{figure}[h]
\centering
\includegraphics[width=8.5cm]{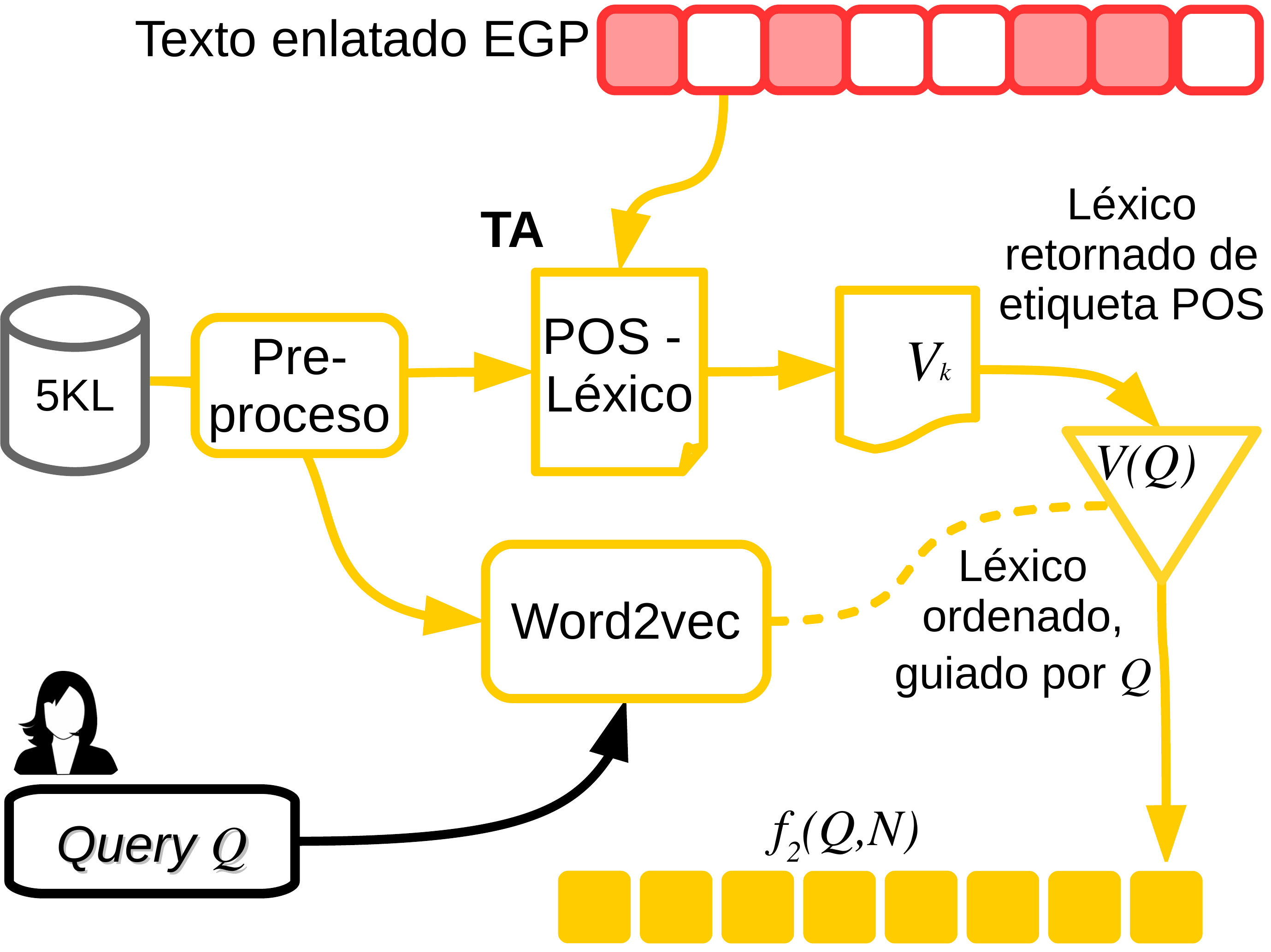}
\caption{Modelo 2: Aproximación semántica basada en \textit{Deep Learning} y análisis morfosintáctico.}
\label{fig:ms2}
\end{figure}

El resultado es una nueva frase $f_2(Q,N)$ que no existe en los corpora 5KL y 8KF.
El proceso se ilustra en la figura \ref{fig:ms2}.

\subsection{Modelo 3: \textrm{Texto enlatado}, aprendizaje profundo e interpretaci\'on geométrica} 
\label{modeloword2vect-3}

El Modelo 3 reutiliza varios de los recursos anteriores: el algoritmo Word2vec, la Tabla Asociativa TA y la estructura gramatical parcialmente vac\'ia (EGP) obtenida del modelo de \textrm{Texto enlatado}.
El modelo utiliza distancias vectoriales para determinar las palabras más adecuadas que sustituirán las etiquetas POS de una EGP y as\'i generar una nueva frase.
Para cada etiqueta POS$_k$, $k=1,2,...$ $\in$ EGP, que se desea sustituir, usamos el algoritmo descrito a continuaci\'on.

Se construye un vector para cada una de las tres palabras siguientes:
\begin{itemize}
    \item $o$: es la palabra $k$ de la frase $f_{o}$  (Sección \ref{sec:modelolata}), correspondiente a la etiqueta POS$_k$. 
    Esta palabra permite recrear un contexto del cual la nueva frase debe alejarse, evitando producir una par\'afrasis.
    \item $Q$: palabra que define al \textit{query} proporcionado por el usuario.
    \item $w$: palabra candidata que podría reemplazar POS$_k$, $w \in V_k$. 
    El vocabulario posee un tamaño $|V_k| = m$ palabras y es recuperado de la TA correspondiente a la POS$_k$.
\end{itemize}

Las 10 palabras $o_i$ más próximas a $o$, las 10 palabras $Q_i$  más próximas a $Q$ y las 10 palabras $w_i$ más próximas a $w$  (en este orden y obtenidas con Word2vec), son concatenadas y representadas en un vector simb\'olico $\vec{U}$ de 30 dimensiones.
El n\'umero de dimensiones fue fijado a 30 de manera emp\'irica, como un compromiso razonable entre diversidad léxica y tiempo de procesamiento.
El vector $\vec{U}$ puede ser escrito como: 
\begin{equation}
\vec{U} = [u_{1},...,u_{10},u_{11},...,u_{20},u_{21},...,u_{30}] \, , 
\end{equation}
\noindent donde cada elemento $u_j, j=1,...,10$, representa una palabra próxima a $o$; $u_j, j=11,...,20$, representa una palabra próxima a $Q$; y $u_j, j=21,...,30$, es una palabra pr\'oxima a $w$.
$\vec{U}$ puede ser re-escrito de la siguiente manera (ecuación \ref{eq:u}): 
\begin{equation}
\label{eq:u}
\vec{U} = [o_{1},...,o_{10},Q_{11},...,Q_{20},w_{21},...,w_{30}]\, .
\end{equation}

\noindent $o$, $Q$ y $w$ generan respectivamente tres vectores numéricos de 30 dimensiones:
\begin{eqnarray} \nonumber
o: \vec X &=& [x_1,...,x_{10},x_{11},...,x_{20},x_{21},...,x_{30} ]\\ \nonumber
Q: \vec Q &=& [q_1,...,q_{10},q_{11},...,q_{20},q_{21},...,q_{30} ]\\ \nonumber
w: \vec W &=& [w_1,...,w_{10},w_{11},...,w_{20},w_{21},...,w_{30} ] \nonumber
\end{eqnarray}
\noindent donde los valores de $\vec X$ son obtenidos tomando la distancia entre la palabra $o$ y cada palabra $u_j \in \vec U, j=1,...,30$.
La distancia, $x_j=dist(o,u_j)$ es proporcionada por Word2vec y adem\'as $x_j \in [0,1]$. 
Evidentemente la palabra $o$ estar\'a m\'as pr\'oxima a las 10 primeras palabras $u_j$ que a las restantes.

Un proceso similar permite obtener los valores de $\vec Q$ y $\vec W$ a partir de $Q$ y $w$, respectivamente.
En estos casos, el $query$ $Q$ estar\'a m\'as pr\'oximo a las  palabras $u_j$ en las posiciones $j=11,...,20$ y la palabra candidata $w$ estar\'a m\'as pr\'oxima a las palabras $u_j$ en las posiciones $j=21,...30$.

Enseguida, se calculan las similitudes coseno entre $\vec{Q}$ y $\vec{W}$ (ecuación \ref{eq:cosQW}) y entre $\vec{X}$ y $\vec{W}$ (ecuación \ref{eq:cosXW}). 
Estos valores también est\'an normalizados entre [0,1].
\begin{equation}
\label{eq:cosQW}
\theta = \cos(\vec{Q},\vec{W}) = \frac{\vec{Q} \cdot \vec{W}}{|\vec Q| |\vec W|}
\end{equation}
\begin{equation}
\label{eq:cosXW}
\beta = \cos(\vec{X},\vec{W}) = \frac{\vec{X} \cdot \vec{W}}{|\vec X| |\vec W|}
\end{equation}
El proceso se repite para todas las palabras $w$ del léxico $V_k$.
Esto genera otro conjunto de vectores $\vec{X}, \vec{Q}$ y $\vec{W}$ para los cuales se deberán calcular nuevamente las similitudes.
Al final se obtienen $m$ valores de similitudes $\theta_i$ y $\beta_i$, $ i= 1,..., m$, y se calculan los promedios $\langle \theta \rangle$ y $\langle \beta \rangle$.

El cociente normalizado $\left( \frac{\langle \theta \rangle} {\theta_i} \right)$ indica qué tan grande es la similitud de $\theta_i$ con respecto al promedio $\langle \theta \rangle$ (interpretaci\'on de tipo maximizaci\'on); es decir, que tan 
 próxima se encuentra la palabra candidata $w$ al \textit{query} $Q$.

El cociente normalizado $\left( \frac{\beta_i} {\langle \beta  \rangle} \right)$ indica qué tan reducida es la similitud de $\beta_i$ con respecto a $\langle \beta \rangle$ (interpretaci\'on de tipo minimizaci\'on); es decir, qué tan lejos se encuentra la palabra candidata $w$ de la palabra $o$ de $f_{o}$.

Estas fracciones se obtienen en cada par $(\theta_i, \beta_i)$ y se combinan (minimizaci\'on-maximizaci\'on) para calcular un score $S_i$, seg\'un la ecuación (\ref{eq:norm}):
\begin{equation}
\label{eq:norm}
   S_i = \left( \frac{\langle \theta \rangle}{\theta_i} \right) \cdot \left( \frac{\beta_i}{\langle \beta \rangle} \right)
\end{equation}
Mientras m\'as elevado sea el valor $S_i$, mejor obedece a nuestros objetivos: acercarse al $query$ y alejarse de la sem\'antica original. 

Finalmente ordenamos en forma decreciente la lista de valores de $S_i$ y se escoge, de manera aleatoria, entre los 3 primeros, la palabra candidata $w$ que reemplazará la etiqueta POS$_k$ en cuestión.
El resultado es una nueva frase $f_3(Q,N)$ que no existe en los corpora utilizados para construir el modelo.

\begin{figure}[h]
\centering
\includegraphics[width=9.5cm]{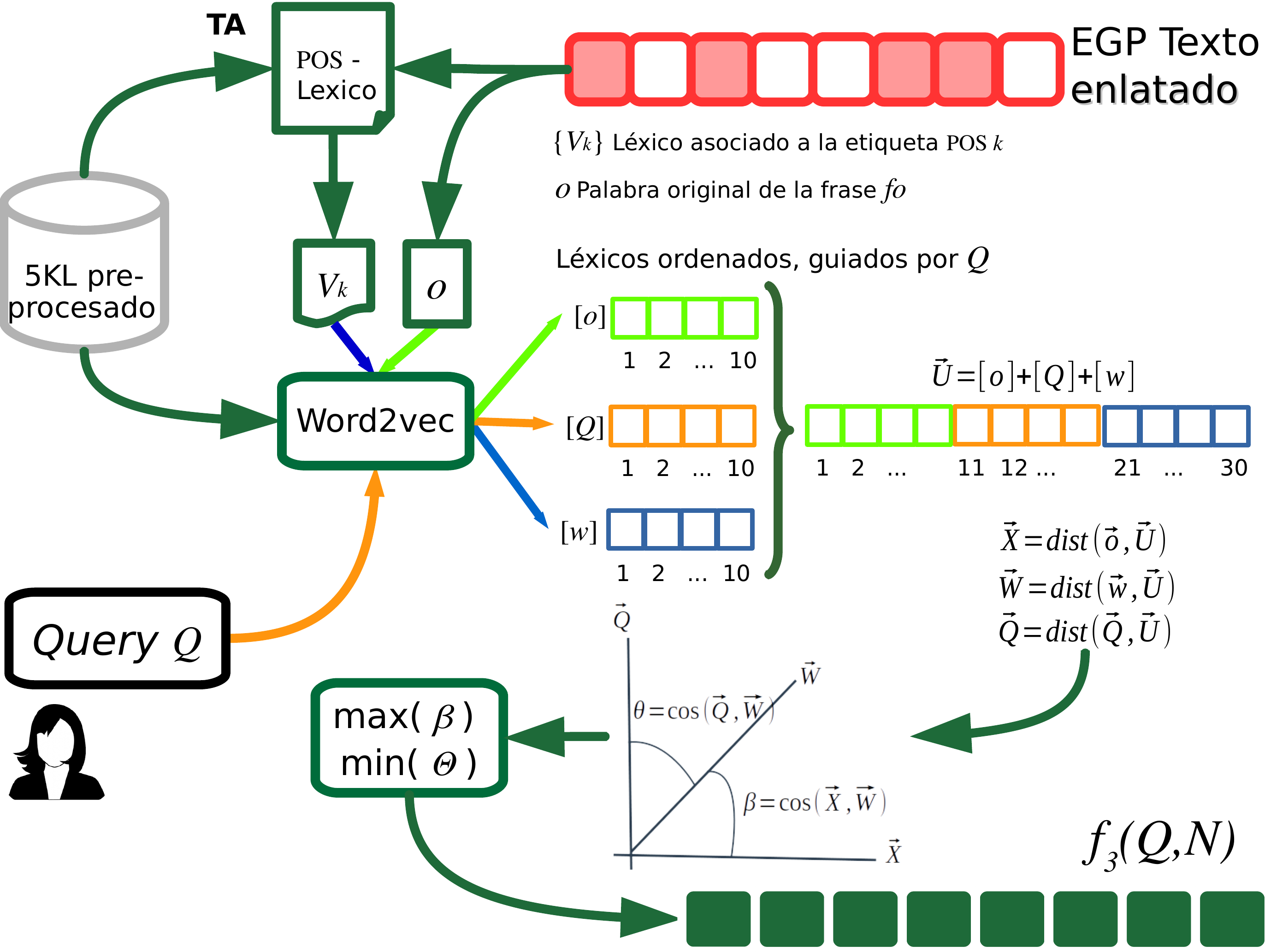}
\caption{Modelo 3: Aproximación semántica basada en interpretación geométrica min-max.}
\label{fig:ms3}
\end{figure}

En la Figura \ref{fig:ms3} se muestra una representación del modelo descrito.

\section{Experimentos y resultados}
\label{sec:resultados}


Dado la especificidad de nuestros experimentos (idioma, corpora disponibles, homosintaxis), no es posible compararse directamente con otros métodos.

Tampoco consideramos la utilizaci\'on de un \textit{baseline} de tipo aleatorio, porque los resultados carecer\'ian de la homosintaxis y ser\'ia sumamente f\'acil obtener mejores resultados.
Dicho lo anterior, el Modelo 1 podr\'ia ser considerado como nuestro propio {\it baseline}.

\subsection{Resultados}

A continuaci\'on presentamos un protocolo de evaluaci\'on manual de los resultados obtenidos.
El experimento consisti\'o en la generaci\'on de 15 frases por cada uno de los tres modelos propuestos. 
Para cada modelo, se consideraron tres \textit{queries}: $Q=$ \texttt{\{AMOR, GUERRA, SOL\}}, generando 5 frases con cada uno. 
Las 15 frases 
fueron mezcladas entre sí y reagrupadas por \textit{queries}, antes de presentarlas a los evaluadores.

Para la evaluaci\'on, se pidi\'o a 7 personas leer cuidadosamente las 45 frases (15 frases por \textit{query}).
Todos los evaluadores poseen estudios universitarios y son hispanohablantes nativos.
Se les pidi\'o anotar en una escala de [0,1,2] (donde 0=mal, 1=aceptable y 2=correcto) los criterios siguientes:

\begin{itemize}
    \item {\bf Gramaticalidad:} ortograf\'ia, conjugaciones correctas, concordancia en g\'enero y n\'umero.
    \item {\bf Coherencia:} legibilidad, percepci\'on de una idea general.
    \item {\bf Contexto:} relaci\'on de la frase con respecto al \textit{query}.
\end{itemize}


Los resultados de la evaluaci\'on se presentan en la Tabla~\ref{tab:eval}, en la forma de promedios normalizados entre [0,1] y de su desviaci\'on estándar $\sigma$.

\begin{table}[h!]
  \centering
  \begin{tabular}{c|rrr}
    \toprule
   \textbf{Modelo} &\bf 1 &\bf 2&\bf 3 \\
    \midrule
        Gramaticalidad  &  0.55     & \it 0.74      & \bf0.77 \\
        $\sigma$      & $\pm$ 0.18      & $\pm$ 0.11  & $\pm$ 0.13\\
         \midrule
        Coherencia      &  0.25     & \it 0.56      & \bf 0.60  \\
        $\sigma$        & $\pm$ 0.15  & $\pm$ 0.11      & $\pm$ 0.14\\
         \midrule
        Contexto        &  \bf0.67  & 0.35          & \it 0.53 \\
        $\sigma$        & $\pm$ 0.25 & $\pm$ 0.17     & $\pm$ 0.19\\
    \bottomrule
  \end{tabular}
  \caption{Resultados de la evaluaci\'on manual.}
  \label{tab:eval}
\end{table}

Las frases generadas por los modelos propuestos presentan características particulares.

El Modelo 1 produce generalmente frases con un contexto estrechamente relacionado con el \textit{query} del usuario, pero a menudo carecen de coherencia y gramaticalidad. 
Este modelo presenta el valor m\'as alto para el contexto, pero también la desviaci\'on est\'andar m\'as elevada.
Se puede inferir que existe cierta discrepancia entre los evaluadores.
Los valores altos para el contexto se explican por el grado de libertad de la EGV generada por el modelo de Markov. 
La EGV permite que todos los elementos de la estructura puedan ser sustituidos por un léxico guiado únicamente por los resultados del algoritmo Word2vec.

El Modelo 2 genera frases razonablemente coherentes y gramaticalmente correctas, pero en ocasiones el contexto se encuentra m\'as pr\'oximo a la frase original que al \textit{query}.
Esto puede ser interpretado como una par\'afrasis elemental, que no es lo que deseamos.

Finalmente, el Modelo 3 genera frases coherentes, gramaticalmente correctas y mejor relacionadas al \textit{query} que el Modelo 2.
Esto se logra siguiendo una intuici\'on opuesta a la par\'afrasis:
buscamos conservar la estructura sintáctica de la frase  original, generando una sem\'antica completamente diferente.

Por otro lado, la mínima dispersión se observa en el Modelo 1, es decir, hay una gran concordancia entre las percepciones de los evaluadores para este criterio. 

\section{Conclusión y trabajo futuro}
\label{sec:conclusiones}

En este art\'iculo hemos presentado tres modelos de producci\'on de frases literarias.
La generaci\'on de este género textual necesita sistemas espec\'ificos que deben considerar el estilo, la sintaxis y una sem\'antica que no necesariamente respeta la l\'ogica de los documentos de géneros factuales, como el period\'istico, enciclopédico o cient\'ifico. 
Los resultados obtenidos son alentadores para el Modelo 3, utilizando Texto enlatado, aprendizaje profundo y una interpretaci\'on del tipo IR.
El trabajo a futuro necesita la implementaci\'on de m\'odulos para procesar los $queries$ multi-término del usuario. 
También se tiene contemplada la generaci\'on de frases ret\'oricas 
utilizando los modelos aqu\'i propuestos u otros con un enfoque probabil\'istico \cite{charton}. 
%
Los modelos aqu\'i presentados pueden ser enriquecidos a trav\'es de la integraci\'on de otros componentes, como características de una personalidad y/o las emociones \cite{We12, plastino2016fisica, Ed17, Si18}.
Finalmente, un protocolo de evaluaci\'on semi-autom\'atico (y a gran escala) est\'a igualmente previsto.

\section*{Agradecimientos}

Los autores agradecen a Eric SanJuan respecto a las ideas y el concepto de la homosintaxis.

%

\bibliographystyle{plain}
\bibliography{biblio.bib}

\end{document}